\documentclass[10pt,twocolumn,letterpaper]{article}

\usepackage{cvpr}              

\usepackage{graphicx}
\usepackage{amsmath}
\usepackage{amssymb}
\usepackage{booktabs}
\usepackage{algorithm}  
\usepackage{algorithmicx}  
\usepackage{algpseudocode}
\usepackage{mathrsfs}
\usepackage{booktabs,tabularx}
\usepackage{multirow}
\usepackage{bbding}
\usepackage{makecell}
\usepackage[pagebackref,breaklinks,colorlinks]{hyperref}
\usepackage{xcolor}

\usepackage[capitalize]{cleveref}
\crefname{section}{Sec.}{Secs.}
\Crefname{section}{Section}{Sections}
\Crefname{table}{Table}{Tables}
\crefname{table}{Tab.}{Tabs.}

\def\confName{CVPR}
\def\confYear{2022}

\begin{document}
\title{Motion-Aware Transformer For Occluded Person Re-identification}

\author{
    Mi Zhou\textsuperscript{\rm 1},
    Hongye Liu\textsuperscript{\rm 2},
    Zhekun Lv\textsuperscript{\rm 3},
    Wei Hong\textsuperscript{\rm 1},
    Xiai Chen\textsuperscript{\rm 2},\\
    \textsuperscript{\rm 1}NetEase,
    \textsuperscript{\rm 2}China JiLiang University,
    \textsuperscript{\rm 3}Hunan Institute of Engineering\\
    \normalsize{\{zhoumi01, hongwei03\}@corp.netease.com,}
    \normalsize{\{liuhongye1998\}@163.com}\\ 
    \normalsize{\{xachen\}@cjlu.edu.cn},
    \normalsize{\{lzk1113666306\}@gmail.com}\\
}   
 
\maketitle

\begin{abstract}
    Recently, occluded person re-identification(Re-ID) remains a challenging task that people are frequently obscured by other people or obstacles, especially in a crowd massing situation. In this paper, we propose a self-supervised deep learning method to improve the location performance for human parts through occluded person Re-ID. Unlike previous works, we find that motion information derived from the photos of various human postures can help identify major human body components. Firstly, a motion-aware transformer encoder-decoder architecture is designed to obtain keypoints heatmaps and part-segmentation maps. Secondly, an affine transformation module is utilized to acquire motion information from the keypoint detection branch. Then the motion information will support the segmentation branch to achieve refined human part segmentation maps, and effectively divide the human body into reasonable groups. Finally, several cases demonstrate the efficiency of the proposed model in distinguishing different representative parts of the human body, which can avoid the background and occlusion disturbs. Our method consistently achieves state-of-the-art results on several popular datasets, including occluded, partial, and holistic.
\end{abstract}

	\section{Introduction}
	\label{sec:intro}
	Person re-identification(Re-ID) has been an active research field for a long time because of its wide range of practical applications such as autonomous driving, action recognition, and surveillance security~\cite{yang2014salient, liao2015person,zheng2012reidentification,zhang2018robust,zhang2018learning}. With the rapid development of the person Re-ID task, various novel methods have been proposed~\cite{zhu2020identity,yang2020spatial, he2021transreid,dai2021idm,ren2021learning} to precisely locate human body parts and personal belongings.
	Despite the existing methods~\cite{wieczorek2021unreasonable,sharma2021person,zheng2019joint,li2021diverse,ni2021flipreid} having achieved exceptional performance on several benchmarks~\cite{miao2019pose,zheng2015partial,zheng2011person,zheng2015scalable,ristani2016performance}, pedestrian occlusion remains a challenging task. 
	Moreover, current occluded person Re-ID methods are more concentrated on gathering robust features from diverse human parts on pixel-level, which need to employ extra pre-trained human parsing models or hand-craft splitting. 
	
	\begin{figure}[t]
		\includegraphics[width=1\linewidth] {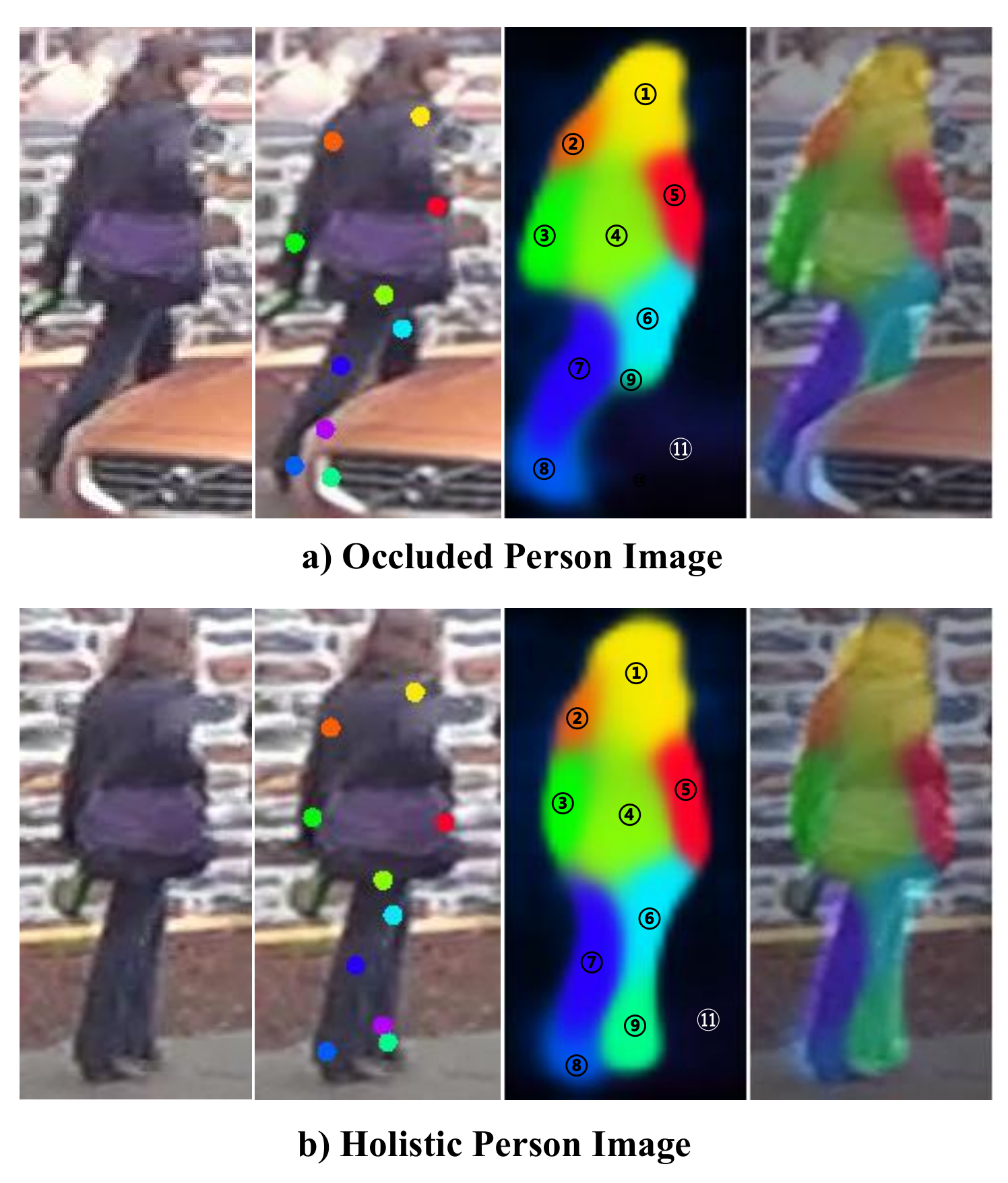}
		\centering
		
		\caption{Visualization of our proposed motion-aware transformer's intermediate results. Original human body images, body images with unsupervised keypoints, part segmentation maps, and body images with part segmentation are shown in sequence on each line. The photos have the same identity in a) and b). When comparing the third image in each line, it is quite easy to spot the identical body component.}
		\label{fig:intro}
	\end{figure}
	
	By reviewing previous occluded person Re-ID methods, we find three key issues are not well tackled. (1) Posture influence. Based on pre-trained human parsing models to achieve pixel-level alignment, methods are always limited by the present human posture, and easy to lose part information, especially in the occluded situation. (2) Lack of the refinement of human body parts. Part-aware attention ~\cite{li2021diverse} and cascaded clustering method ~\cite{zhu2020identity} have proposed obtaining more refined human part features to align person Re-ID better. Nevertheless, the part-aware attention method can not sufficiently take the details of the diverse human parts, but it has brilliant performance on color and texture feature extraction. The cascaded clustering method shows excellent performance in clustering various parts of the human body and personal belongings. However, it still can not eliminate the dilemma when the background and people are too similar. (3) Lack of modeling the relationship of various parts of the human body. Although, local relation information is exploited in several recent  studies~\cite{zhang2021person,zheng2019pose,jiang2019ph,park2020relation}, where they compute relation maps based on the local similarity among feature maps or aggregate local features to learn relation information from different body parts. But they are still reliant on hand-crafted splitting methods that divide the image or feature map into small patches or rigid stripes, which is too coarse to align the human parts well and introduces lots of background noise.
	
	To mitigate the above issues, we propose an effective motion-aware transformer-based architecture specially designed for the difficulties in occluded person Re-ID. Specifically, we first adopt a transformer encoder-decoder for modeling the relationship of the images features from ResNet50~\cite{ResNet}. Then we design an MLP prediction head for two branches: keypoint detector and part segmentation. In the keypoint detector branch, we design it as a self-supervised process to overcome the pose-influence dilemma caused by pre-trained human parsing models and set an affine transformation module to learn motion information. The motion information will help the model to decide which parts of the human body can be clustered into a class. What is more, it is the first work to introduce motion information for person Re-ID. In the part segmentation branch, we utilize the motion information from the keypoint branch to achieve refined human part segmentation and obtain representative local features, as shown in Fig.~\ref{fig:intro}. 
	
	
	Different from previous approaches, we propose a novel motion-aware transformer-based model architecture that performs occluded person Re-ID based on motion information and part segmentation, which has three prominent advantages: 
	\begin{enumerate}
	\item With the motion information combined with part segmentation, we can overcome the complexity posture influence, avoid the disturbance by background noise, as well as model better relationships of different parts of the human body.
	\item The keypoint detector can be trained in a self-supervised manner and optimized mutually to extract motion information better. The part segmentation not only contains representative local features but the correspondence of each human body part in different images, which significantly enhances the accuracy for occluded person Re-ID task.
	\item The transformer model optimization can lead the keypoint detector to locate more precisely and lead the part segmentation to represent more detailed human body information.
	\end{enumerate}
	
	To demonstrate the effectiveness of the proposed motion-aware scheme, we conduct extensive ablation studies and experiments on several popular datasets~\cite{miao2019pose, zheng2015partial, zheng2011person,zheng2015scalable,wei2018person}, consistently achieving the state-of-of-art results. 
	


	\begin{figure*}
		\begin{center}
			\includegraphics[width=1.0\linewidth] {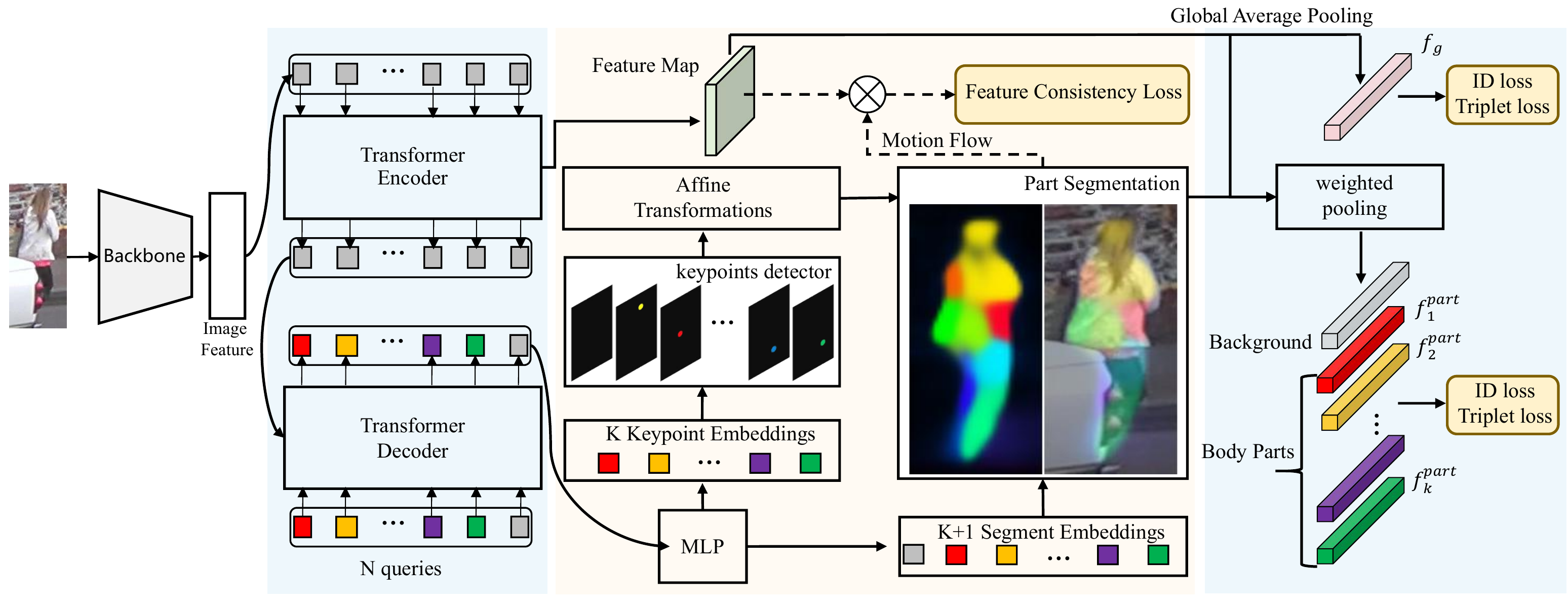}
		\end{center}
		\caption{The pipeline of the proposed motion-aware transformer consists of feature extraction, a motion flow estimator in a self-supervised manner, and a part segmentation module. Following the transformer decoder, there are two branches: a self-supervised keypoint detector for predicting motion flow and a part segmentation module.}
		\label{fig-pipeline}
	\end{figure*}

	\section{Related Work}
	\subsection{Occluded person Re-ID}
	
	Occluded person Re-ID aims to find the same person with holistic or occluded appearances from different camera views~\cite{liang2018look, zheng2016person}. Through early research on feature extraction and metrix learning~\cite{Reidentification,koestinger2012large,liao2015efficient, liao2015person}, researchers have proposed various novel methods to learn representative local features in order to better adapt to complex situations, especially in occlusion situations. Moreover, occluded person Re-ID remains challenging due to incomplete information and spatial misalignment.
	
	With the rapid development of the person Re-ID tasks, plenty of practical approaches have been proposed to alleviate the dilemma of misalignment. According to previous approaches, the fundamental methods can be defined in various forms. (1) Stripe-based methods directly divide the person images into horizontal stripes to obtain their striped-local features~\cite{yi2014deep,zhang2017alignedreid,luo2019alignedreid++,fu2019horizontal}. However, the striped-based methods lack sufficient fine-grained information to align human parts and can not suppress the impact of background introduction. (2) Human parsing-based methods pay more attention to inherent human body features such as body posture and key points to better overcome the background noise disturbance~\cite{zhao2017spindle, zheng2019pose}. However, it excessively depends on pre-trained human parsing models, which produce a fixed number of latent parts and can not tackle the occluded images.
	(3) Human semantic-based methods introduce extra semantic in order to locate body parts and better acquire local features~\cite{guler2018densepose,guo2019beyond,kalayeh2018human,liu2018pose,sarfraz2018pose,song2018mask}. However, with the introduction of segmentation, the utility and robustness of these methods have been limited. Moreover, they can not revise the mistakes throughout the training.
	
	Although the above methods remain many defects, they all bring novel ideas for the occluded Re-ID tasks. AFPB~\cite{zhuo2018occluded} proposed occluded and person ID classification with a multi-task loss to extract representative information. Miao \emph{et al.}~\cite{miao2019pose} propose to utilize posture landmarks to disentangle the valuable information from the occlusion. He\emph{et al.} ~\cite{he2019foreground}
	proposed FPR to utilize the error from robust reconstruction over spatial pyramid features to measure similarities between two persons. PVPM~\cite{gao2020pose} proposed a pose-guided attention to learn discriminative part features. ISP~\cite{zhu2020identity} proposed a pixel-level method with cluster assignments to locate both the human body parts and personal belongings. 
	
	\subsection{Transformer-based methods for person Re-ID}
	Transformer-based methods~\cite{transformer, dosovitskiy2020image}  achieve state-of-the-art performance on several image processing tasks. Research shows that Transformer compared with the CNN method, has its unique superiority on global feature learning and representation. Based on Transformer,  PAT~\cite{li2021diverse} has firstly proposed an end-to-end Part-Aware Transformer for occluded person Re-ID by using identity labels only and acquiring favorable performance. TransReID~\cite{he2021transreid} has proposed a pure transformer-based object Re-ID framework with jigsaw patch module and side information embeddings.

	\subsection{Motion Description}
	Image animations have gained popularity as they are key-enabling components for automatic video editing. Aliaksandr \emph{et al.}~\cite{siarohin2019first} proposed the local affine transformation for approximate motion description. Specifically, the first order Taylor expansion has been used in a neighborhood of the keypoint locations to describe the approximate motion. Then a generation module has been utilized to a generation network to reconstruct the target. These works significantly demonstrate that motion information can and should be adopted for inferring meaningful object parts. Motion-supervised co-part segmentation~\cite{siarohin2021motion}, the pioneering work proposed a novel architecture via an approach of self-supervised and reconstruction for co-part segmentation. It constitutes intermediate motion representations robust to sensor changes and appearance variations, as well as eliminates the background disturbance. As for the person Re-ID task, designing a generation or reconstruction module is difficult due to the discreteness of person images, but what if we design a motion-aware module without a reconstruction or generation module?
	
	To address the above issues, we present a creative motion-aware transformer-based architecture for occluded person Re-ID, which considers motion information by a self-supervised process to improve the robustness to sensor changes and appearance variations. Secondly, we utilize the transformer feature aggregation capabilities with global-range and long-distance dependencies to better extract local and global features. Thirdly, we design a new loss to replace the complicated entire generation module and make the two branches, self-supervised keypoint detection and part-segmentation, information blend with each other. Our method has performed consistently on several popular benchmarks for person Re-ID.

	\section{Methodology}
	\label{sec:method}
	We first introduce the overall architecture of the proposed occluded person Re-ID method. Then we elaborate on our proposed motion-aware transformer while analyzing the advantages of introducing motion information. Finally, we describe the details about the training and inference of our proposed method. 
	
	\subsection{Network Architecture}
	\label{sec:architecture}
	As shown in Fig.~\ref{fig-pipeline}, our proposed motion-aware transformer-based architecture for occluded person Re-ID is composed of a feature extractor, a self-supervised motion flow estimator, and a part segmentation branch. 
	
	Firstly, we use a CNN backbone to extract the features $\mathcal{F} \in \mathbb{R}^{h \times w \times D}$ of each human image $I \in \mathbb{R}^{H \times W \times C}$, and then we flatten the output features into one dimension as the input for the transformer encoder, where $\mathcal{F}$ is in a $hw \times D$ feature. Before being fed into the transformer encoder, the feature maps are first concatenated with position embeddings. Then we adopt an alternative transformer to model the relationship, implicitly aggregating the features from the same person and finally obtain global feature maps $f_g \in \mathbb{R}^{h \times w \times d_g}$. Two branches follow the transformer decoder for motion flow estimation and part segmentation, from which we can obtain $K$ keypoint embeddings and $K+1$ segment embeddings. Then, we multiply the feature maps and the embeddings to obtain keypoint heatmaps and segmentation maps, respectively. Motion flow estimator based on keypoint heatmaps is designed in a self-supervised manner to acquire motion information, through which we obtain segments that group pixels corresponding to body parts that move together. Each keypoint heatmap corresponds to a body part, and the background region is predicted using an extra segment. Finally, we obtain $K+1$ refined human part segmentation maps and utilize the weighted pooling to get representative body component features $f^{part} \in \mathbb{R}^{h \times w \times d_p}$.


	\subsection{Feature Extraction}
	Background areas with a variety of characteristics make obtaining robust features for the target person more challenging. Therefore, we use a transformer encoder to capture the entire image context information. Our method uses ResNet-50~\cite{ResNet} without the average pooling layer and fully connected layer as the backbone to extract global feature maps from specific images. After the backbone network extracts the image features, we adopt a transformer structure to implicitly aggregate features from the same person and further model the relationship of different instances.
	The basic form is a stacked network with six transformer encoder layers and a single layer of the decoder part, which are composed of self-attention layers, feed-forward networks, and layer normalization. In each self-attention operation, we have:
	\begin{equation}
		Attn(\widetilde{\mathcal{F}}) = softmax(\frac{\widetilde{\mathcal{F}}\widetilde{\mathcal{F}}^T}{\sqrt{D'}})\widetilde{\mathcal{F}}
	\end{equation}
	
	
	where $\widetilde{\mathcal{F}} \in \mathbb{R}^{hw \times D'}$ is the input human feature with the position embeddings, and here $\frac{1}{\sqrt{D'}}$ is the scaling fator for normalization.
	
	
	Finally, the output person features are fed into two prediction heads for keypoint detection and part segmentation.  These two prediction heads composed of MLP with a sigmoid function generate the $K$ keypoint embeddings and $K+1$ part segmentation embeddings.
	
	\subsection{Motion Flow Estimator}
	\label{sec:self-supervised}
	The motion flow estimator aims to predict a dense motion field from a referred body image to a target body image with the same identity. Inspired by previous approaches~\cite{siarohin2019first, siarohin2021motion}, the motion estimator module proceeds in two steps. 	Firstly, we approximate both transformations from sets of sparse trajectories, obtained by using keypoints learned in a self-supervised manner. Secondly, we utilize a dense motion network combining the local approximations to obtain the resulting dense motion field.
	
	
	\textbf{Self-supervised keypoint detector} As shown in our pipeline~\ref{fig-pipeline}, we utilize an MLP head followed by the transformed decoder and a keypoint detector module to predict keypoints. The most pressing problem is determining how to effectively restrict the whole self-supervised process during training and achieving the intended impact. However, our keypoint detector does not need any annotations, we utilize a normal self-supervised equivariance loss~\cite{siarohin2019first} to make directly constraint.
	
	Obviously, compared with current methods, the self-supervised method does not rely on any additional prior information, pre-trained model as well as any extra annotations. 
	
	\textbf{Local Motion Description} The motion estimation module estimates the backward optical flow from a reference body image $\mathcal{I}_r$ to the target body image $\mathcal{I}_t$. After obtaining $K$ keypoint embeddings from the feature maps $F_m$ with MLP layer, we will obtain:
	\begin{equation}
		\begin{aligned}
			&P &= & \{z_i \in F_m~|~i = 0,1,...,k\}\\
		\end{aligned}
	\end{equation} 
	where $z_1,z_2,...,z_k$ represents coordinates of the keypoints in feature maps. 
	
	Following~\cite{siarohin2019first}, we assume that there exists an standard abstract image $\mathcal{O}$ and we firstly consider the most general case that all keypoints are not missing. Thus, we easily obtain that estimating $A_{\mathcal{I}_r \gets \mathcal{I}_t}$ consists in estimating $A_{\mathcal{I}_r \gets \mathcal{O}}$ and $A_{\mathcal{O} \gets \mathcal{I}_t}$. Formally, we consider the first order Taylor expansions in $K$ keypoints and calculate its Jacobians matrix $J_k$ in each $p_k$ location. Finally we obtain:
	\begin{equation}
		\mathcal{A}_{\mathcal{I}_r \gets \mathcal{I}_t}(z) \approx \mathcal{A}_{\mathcal{I} \gets \mathcal{O}}(p_k) + J_k(z-\mathcal{A}_{\mathcal{O} \gets \mathcal{I}_t}(p_k))
	\end{equation}
	with:
	\begin{equation}
		\small
		J_k = \left(\frac{d}{dp}\mathcal{A}_{\mathcal{I}_r \gets \mathcal{O}}(p)\Bigg|p=p_k\right)\left(\frac{d}{dp}\mathcal{A}_{\mathcal{I}_t \gets \mathcal{O}}(p)\Bigg|p=p_k\right)^{-1}
		\label{JK}
	\end{equation}
	where $z$ represents the point locations in $\mathcal{I}_r$ and $\mathcal{I}_t$ and $p$ denotes the coordinates of the keypoints in the standard image $\mathcal{O}$. 
	
	
	\subsection{Part Segmentation}
	We now detail how we model the motion of each segment. We design our model to obtain segments that group pixels corresponding to body parts that move together and utilize an affine transformation to model the motion of the pixels within each body part. So we can use motion flow to support body part segmentation.  In keypoint detecting branch, we obtain the $K$ keypoint heatmaps and corresponding the Jacobians. Then imagine that we have two body images from the same people with different postures. The one is the reference image $\mathcal{I}_r$, and the other one is the target image $\mathcal{I}_t$. We utilize $M_{\mathcal{I}}^k$ to represent the $k+1$ channel of the part segmentation $M_{\mathcal{I}}$. Then we define $M_{\mathcal{I}_r}^k \in [0,1]$ as the set of locations of the reference image part segment $k$ and $M_{\mathcal{I}_t}^k \in [0,1]$ as the set of locations of the target image part segment $k$. To be more specific, the critical task here is to acquire the motion flow $F$ from segmentation maps and affine transformation. From the self-supervised keypoint detector, we have obtained an affine transformation of each segment, and then we can formulate the motion flow as:
	\begin{equation}
		F(z) = s_{\mathcal{I}_r}^k + J_k(z-s_{\mathcal{I}_t}^k), z\in M_{\mathcal{I}_t}^k
		\label{flow}
	\end{equation}
	
	where $s_{\mathcal{I}_r}^k, s_{\mathcal{I}_t}^k \in \mathcal{R}^2$ represents the pixel location corresponding to the segmentation respectively and $z$ denotes the point locations in $M_{\mathcal{I}_t}$ pose spaces. Then for each keypoint we can acquire:
	
	\begin{equation}
		\mathcal{M}_{\mathcal{I}_{rt}}(p_k) = \mathcal{A}_{\mathcal{I}_r(p^r_k) \gets \mathcal{I}_t(p^t_k)}, k \in [1, K]
	\end{equation}
	where $\mathcal{M}$ represents the affine transformation among images. With Eq.~\ref{flow}, we can utilize the entire image feature maps and the affine transformation to calculate the motion flow as:
	\begin{equation}
		F_{r \gets t}(z) = s_{\mathcal{I}_r}^k + \mathcal{M}_{\mathcal{I}_{rt}}(z-s_{\mathcal{I}_t}^k)
	\end{equation}
	where $F_{r \gets t}$ denote the motion flow between image $\mathcal{I}_r$ and $\mathcal{I}_t$. Additionally, we set $f^r,f^t$ as the reference and target image features respectively, then we can acquire a constraint as:
	\begin{equation}
		f^r \equiv f^{rt} = F_{r \gets t} \times f^t
		\label{hard eq}
	\end{equation}
	
	\subsection{Optimization}
	The proposed model can be trained in an end-to-end manner, and the loss function can be formulated as follows:
	\begin{equation}
		\mathcal{L} = \lambda_1 \mathcal{L}_{id}^g + \lambda_2 \mathcal{L}_{t}^g + \lambda_3 \mathcal{L}_{id}^p + \lambda_4 \mathcal{L}_{t}^p + \lambda_5 \mathcal{L}_{eq}+\lambda_6 \mathcal{L}_{fc}
	\end{equation}
	
	where $\mathcal{L}_{id}^g$ and $\mathcal{L}_{id}^p$ denote the loss for classification based on global features and local features respectively, $\mathcal{L}_{t}^g$ and $\mathcal{L}_{t}^p$ denote the triplet loss for feature embeddings based on global features and local features respectively. $\mathcal{L}_{eq}$ denotes the loss for constraining the self-supervised process and stabilize the entire training process, and $\mathcal{L}_{fc}$ denotes the loss for optimizing part segmentation. Here, we set $\lambda_1=\lambda_2=\lambda_3=\lambda_4=1$, $\lambda_5=10$ and $\lambda_6=5$.
	
	\textbf{Feature Consistency loss} In the training stage, we will choose four images $\{\mathcal{I}_A, \mathcal{I}_B, \mathcal{I}_C, \mathcal{I}_D\}$ with the same ID. Then through keypoint detector, we will obtain their keypoints $\{p^A_k, p^B_k, p^C_k, p^D_k| k \in [1, K]\}$.Note that in the training stage, Eq.~\ref{hard eq} will hardly be equal, but our purpose is to let the above equation hold. Formally, if we can acquire the motion flow between two images, we can use them to transform the two distributions $f^A$ and $f^B$ into each other. Therefore, we utilize the Kullback–Leibler divergence, simply KL, to constrain this equation as:
	\begin{equation}
		D_{kl}(f^A \parallel f^{AB}) = \sum_zf^A(z)\log \frac{f^A(z)}{f^{AB}(z)}
	\end{equation}
	where $D_{kl}$ represents the KL divergence. When we consider the situation for four images, we utilize $f^A$ as the referenced image feature. Therefore we can easily acquire $F_{A \gets B}, F_{A \gets C}, F_{A \gets D}$ , then we can obtain:
	\begin{equation}
		\left\{
		\begin{aligned}
			f^{AB} &=& F_{A  \gets B} \times f^B\\
			f^{AC} &=& F_{A  \gets C} \times f^C\\
			f^{AD} &=& F_{A  \gets D} \times f^D\\
		\end{aligned}
		\right.
	\end{equation}
	With above equation, the final feature consistency loss can be formulated as:
	\begin{equation}
		\small
		\mathcal{L}_{fc} = D_{kl}(f^A \parallel f^{AB}) + D_{kl}(f^A \parallel f^{AC}) + D_{kl}(f^A \parallel f^{AD})
	\end{equation}
	
	\textbf{Equivariance loss} The primary equivariance constraint is one of the essential factors driving the discovery of unsupervised keypoints~\cite{siarohin2019first, jakab2018unsupervised, zhang2018unsupervised}, which will force the model to predict consistent keypoints concerning known geometric transformations and stabilize the training process. Our model predicts not only the keypoints but also the Jacobians, and the standard equivariance constraint can be formulated as:
	\begin{equation}
		\mathcal{A}_{\mathcal{X}\gets \mathcal{O}} = \mathcal{A}_{\mathcal{X}\gets \mathcal{Y}} \circ \mathcal{A}_{\mathcal{Y}\gets \mathcal{O}}
	\end{equation}
	where $\mathcal{O}$ represents the abstract standard images with entire keypoints, $\mathcal{X}$ represents the original image, and $\mathcal{Y}$ represents the transformed image. Significantly, we assume that the keypoint location in $\mathcal{O}$ is already known and will not change during the whole process, and $A_{\mathcal{X}\gets \mathcal{Y}}$ denote the known thin-plate splines deformations transformation. Finally, we take Jacobians into consideration and adopt the $L_1$ loss to constrain every keypoint locations and the Jacobians, and we can obtain:
	\begin{equation}
		\begin{aligned}
			\mathcal{L}_{eq} = &\parallel	\mathcal{A}_{\mathcal{X}\gets \mathcal{O}} - \mathcal{A}_{\mathcal{X}\gets \mathcal{Y}} \circ \mathcal{A}_{\mathcal{Y}\gets \mathcal{O}} \parallel_1 + \\
			&\parallel J_k\left(\frac{d}{dp}\mathcal{A}_{\mathcal{X}\gets \mathcal{Y}}(p) \Big|_{p=\mathcal{A}_{\mathcal{Y} \gets \mathcal{O}(p_k)}}\right) - \mathbb{E} \parallel_1
		\end{aligned}
	\end{equation}
	
	where $J_k$ is as same as in Eq.~\ref{JK} and $\mathbb{E}$ represents the $2 \times 2$ identity matrix. 
	
	\textbf{Identity loss} For global branch, we have acquire global features $f^g$, then we use $softmax$ to obtain its classification result, we can obtain:
	\begin{equation}
		\mathcal{L}_{id}^g = -\frac{1}{N}\sum_{i=1}^N\log(p(y_i|x_i))
	\end{equation}
	where $N$ represents the samples, $p(y_i|x_i)$ represents the predicted probability that $x_i$ is recognized as $y_i$. The local branch is as same as the global branch:
	\begin{equation}
		\mathcal{L}_{id}^p = -\frac{1}{N}\sum_{i=1}^N\log(p(y_i|x_i))
	\end{equation}
	
	\textbf{Triplet loss.} Triplet loss~\cite{schroff2015facenet} is first proposed to learn the embedding of the human face better. Then Triplet loss has been utilized in many deep learning fields as a typical loss function aiming to make the same sample features more closely in embedding space and make different features as far as possible. For a triplet set $\{a,p,n\}$, the triplet loss with soft-margin is formulated as follows:
	\begin{equation}
		\small
		\mathcal{L}_{t}^g=\mathcal{L}_{p}^g=\log\left[1+exp(\left( \parallel  f_a - f_p \parallel_2^2 - \parallel  f_a - f_n \parallel_2^2 \right)\right]
	\end{equation}

	\section{Experiments}
	\label{sec:exp}
	In this section, we firstly verify the performance of our proposed model for occluded person Re-ID, partial Re-ID and holistic Re-ID. Then adequate ablation studies will be performed to prove the effectiveness of each module. 
	
	\subsection{Datasets}
	We conduct extensive experiments on two occluded person Re-ID datasets: Occluded REID~\cite{miao2019pose} and Occluded Duke~\cite{miao2019pose}, two partial Re-ID datasets: Partial REID~\cite{zheng2015partial} and Partial-iLIDS~\cite{zheng2011person} and three holistic Re-ID datasets: Market-1501~\cite{zheng2015scalable}, DukeMTMC-reID~\cite{ristani2016performance} and MSMT17~\cite{wei2018person}. The detailed description of the above datasets are as follows:
	
	\textbf{Occluded REID}~\cite{miao2019pose} is an occluded person Re-ID dataset which is captured by the mobile cameras. It has 2,000 images belonging to 200 identities, and each identify has five full-body images and five occluded images with different types of occlusions. All images are resized to 128 $\times$ 64.
	
	\textbf{Occluded Duke}~\cite{miao2019pose} has 15,168 training images, 17,661 gallery images and 2,210 occluded query images which is designed for the occluded person re-id problem. These images are re-splited from the DukeMTMC-reID dataset by keep occluded images and filtering out some overlap images.
	
	\textbf{Partial REID}~\cite{zheng2015partial} is a benchmark for partial person Re-ID task which contains 600 images from 60 people with five full-body images in gallery set and five partial images in query set per person.	
	
	\textbf{Partial-iLIDS}~\cite{zheng2011person} is a partial person Re-ID simulation dataset based on iLIDS~\cite{zheng2011person}. Partial-iLIDS contains 238 images from 119 people captured by multiple cameras in the
	airport, and their occluded regions are manually cropped.
	
	\textbf{Market-1501}~\cite{zheng2015scalable} contains 1,501 identities captured by 6 cameras. The training set contains 12,936 images of 751 identities, the query set contains 3,368 images and the gallery set contains 19,732 images.
	
	\textbf{DukeMTMC-reID}~\cite{ristani2016performance} contains 36,411 images from 1,404 identities captured by 8 cameras. The training set contains 16,522 images, the query set contains 2,228 images and the gallery set contains 17,661 images.
	
	\textbf{MSMT17}~\cite{wei2018person} contains 126,441 images from 4,101 identities captured by 15 cameras. The training set contains 32,621 images of 1,041 identities, the query set contains 11,659 images and the gallery set contains 82,161 images of 3,060 identities.
	
	\textbf{Evaluation Metrics} We adopt the mean Average Precision (mAP) accuracy and the Cumulative Matching Characteristic (CMC) curve at Rank-1 accuracy and Rank-5 accuracy as the evaluation protocols to assess the performance of our proposed model.

	\subsection{Implementation Details}
	We use ResNet-50~\cite{ResNet} as our backbone, and all the person images are resized to $256\times128$. When in the training stage, we utilize random horizontal flipping, padding, random cropping, and random erasing as data augmentation. The models are trained on 8 RTX-3090 GPUs for 100 epochs, and batch size is set to $64$ with four same ID images. We optimize our model by an SGD optimizer with a momentum of 0.9 and the weight decay of $1e^{-4}$. The learning rate we initially set is 0.01 and will decay $0.1$ for every $30$ epoch.  
	
	In the inference period, we keep resizing images to $256\times128$, and the inference is testing on a single RTX-3090 with a batch size of 1.

	\begin{table*}
		\centering
		\setlength\tabcolsep{4pt}
		\caption{Performance comparison with different key modules on Occluded Duke}
		\renewcommand\arraystretch{1}
		\begin{tabularx}{0.95\linewidth}{cccccccccc}
			\toprule
			\multirow{2}*{Index} &\multirow{2}*{Encoder}& \multirow{2}*{Decoder}& \multirow{2}*{Segmentation Branch} & \multicolumn{2}{c}{Keypoint Branch}& \multirow{2}*{Rank-1} & \multirow{2}*{Rank-5} &\multirow{2}*{Rank-10} & \multirow{2}*{mAP} \\
			~&~&~&~&Pre-train&Self-supervised&~&~&~&~\\
			\midrule
			\#1 &  &  &  & &  &46.0 &65.7 &71.9 &38.8 \\
			\#2 & &   &\Checkmark  & &\Checkmark &63.3 &78.6 &83.3 &54.7 \\
			\#3 &\Checkmark &   &\Checkmark &&\Checkmark &65.9 &80.1 &85.2 &57.3 \\
			\#4 & &\Checkmark   &\Checkmark &&\Checkmark	&64.7 &79.8 &84.7 &56.9 \\
			\#5 &\Checkmark &\Checkmark &\Checkmark & &\Checkmark &\textbf{66.2} &\textbf{80.8} &\textbf{85.6} &\textbf{58.8} \\
			\#6 &\Checkmark &\Checkmark &\Checkmark &\Checkmark& &64.2 &78.5 &83.6 &54.3\\
			\bottomrule	\end{tabularx}
		\label{tab:module}
	\end{table*}

	\subsection{Ablation Experiments}
	To demonstrate the effectiveness of our proposed method, we conduct comprehensive experiments with each module configuration on Occluded Duke~\cite{miao2019pose}. 
	
	In this part, we set ResNet50 as our baseline. As shown in the rows \#1,\#2 of the Tab.~\ref{tab:module}, the method based on segmentation branch with self-supervised keypoint detector improves Rank-1 accuracy from $46.0\%$ to $63.3\%$ and mAP from $38.8\%$ to $54.7\%$. In the rows \#2,\#3, we can conclude that the transformer encoder has the powerful ability to aggregate the valuable features, demonstrating the superiority of stacking self-attention layers, which make the Rank-1 accuracy $+2.6\%$ and mAP $+2.6\%$. 
	
	In the rows \#3,\#4 we can conclude that the transformer decoder can bring a certain degree of improvement, but the encoder part makes more contribution than the decoder part, which also explains the reason why we utilize the feature maps through the encoder as our global features. In the rows \#3,\#5 the combine of the transformer encoder and decoder achieves $66.2\%$ and $58.8\%$ compared with encoder only on $65.9\%$ and $57.3\%$. In the rows \#5,\#6 we find that When we utilize the self-supervised keypoint detector, it significantly improves Rank-1 accuracy and mAP by $2.0\%$ and $4.5\%$ compared with the pre-train detector $64.2\%$ and $54.3\%$, which strongly demonstrate the effectiveness of the self-supervised detector. 
		
	In conclusion, this ablation study demonstrates that: 1) the combination of self-supervised keypoint detector and part segmentation make the most contribution. 2) the transformer encoder and decoder is not the decisive factor to bring improvement. In other words, the transformer just provides a better ability on feature aggregation.

	\subsection{Analysis of the Number of part Segmentation}
	To analysis the influence of the part number for our proposed part segmentation scheme, we conduct several experiments with different part number configurations on Occluded Duke~\cite{miao2019pose}. 
	
	As shown in Tab.~\ref{tab:part}, we compare the impact of the different numbers of segmentation parts on the performance of the model. We can find that too little or too much part number will directly influence the performance. When part numbers are too few, it is approximate to ISP~\cite{zhu2020identity}, which can not fully overcome the dilemma brought by the posture occlusion. When part numbers are too much, it becomes unstable, more sensitive, and brings more background noise, significantly decreasing the performance.

	Specifically, when the part number increases from 3 to 10, the Rank-1 accuracy and mAP have significant improvement from $64.4\%, 55.3\%$ to $66.2\%, 58.8\%$, and reach the peak at part number 10. However, when the part number continues to increase to 17, the Rank-1 accuracy and mAP decrease from $66.2\%, 58.8\%$ to $65.0\%, 56.1\%$, whose performance is so close to part number 6.

\begin{figure}[t]
		\begin{center}
			\includegraphics[width=1.0\linewidth] {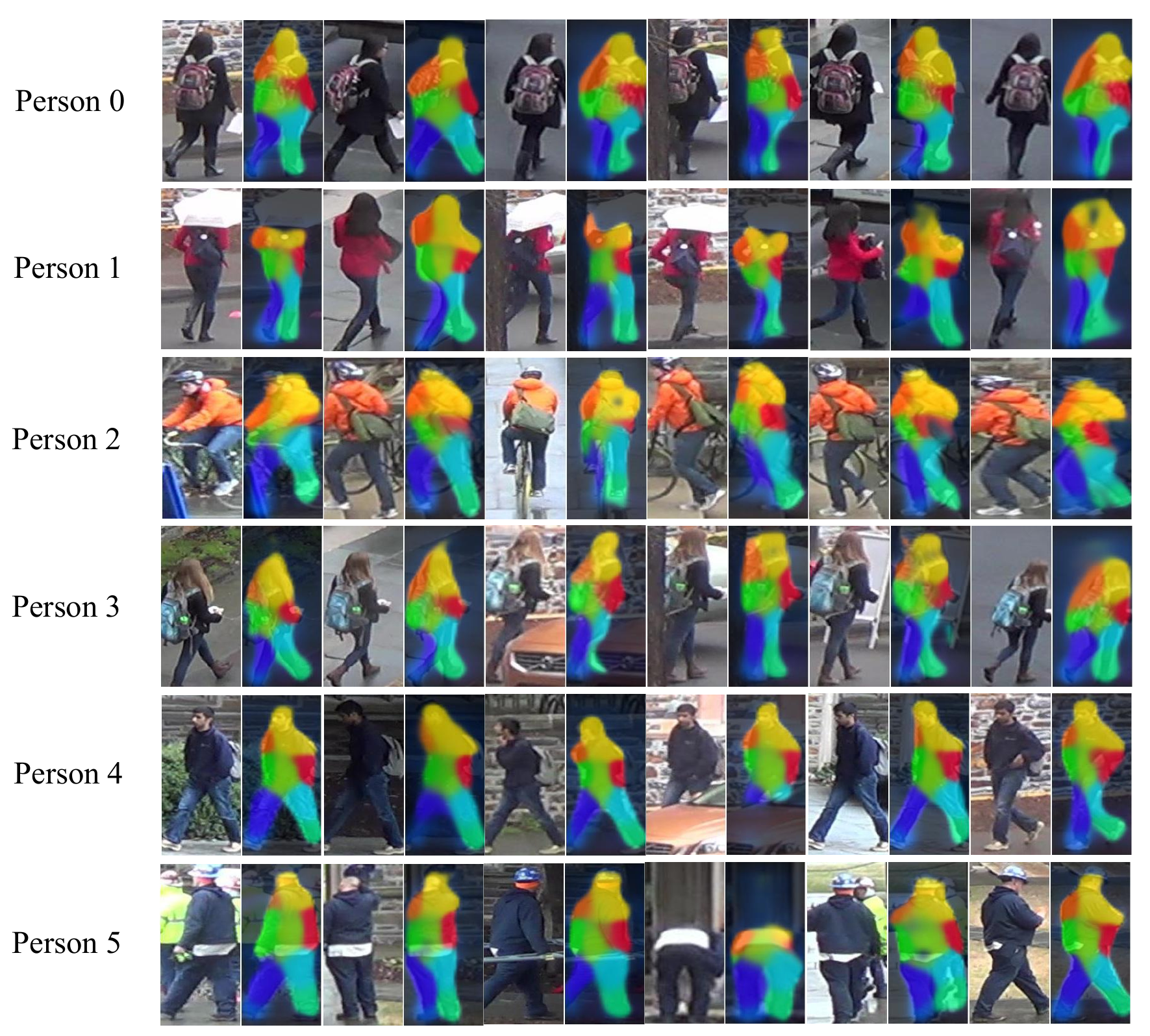}
		\end{center}
		\caption{Visualization of the learned part segmentation. Each line shows the body images of the same people with different postures. As we can see, these part segmentations mainly focus on different discriminative human parts.}
		\label{fig:visual}
	\end{figure}

	Above all, part numbers of part segmentation is an essential and sensitive parameter that could directly influence the model performance. Note that too little or too much both have impact on the final performance. Consequently, through our experiments, ten parts are the best number of part segmentation branches. As shown in Fig.~\ref{fig:visual}, we can find that our method could achieve refined human part segmentation to support the local feature extraction. 
		
	\begin{table}[t]
		\centering
		\setlength\tabcolsep{4.2pt}
		\renewcommand\arraystretch{1}
		\caption{Different part number of part segmentation}
		\begin{tabularx}{1\linewidth}{c|c|cccc}
			\hline
			Index &part number& Rank-1 & Rank-5&Rank-10 & mAP \\
			\hline
			\#1& 3&64.4 &79.4 &84.1 &55.3 \\
			\#2& 6&65.2 &79.7 &84.3 &56.2 \\
			\#3& 10&\textbf{66.2} &\textbf{80.8} &\textbf{85.6} &\textbf{58.8} \\
			\#4& 17&65.0 &79.5 &84.2 &56.1 \\
			\hline
		\end{tabularx}
		\label{tab:part}
	\end{table}

	\begin{table*}
		\centering
		\small
		\setlength\tabcolsep{1.1pt}
		\renewcommand\arraystretch{1}
		\caption{Performance comparison with state-of-the-arts on Occluded-Duke dataset, Occluded-REID dataset, Partial REID, Partial-iLIDS, Market-1501 and DukeMTMC-reID datasets. Our method achieves the best performance on five datasets.}
		\begin{tabularx}{1\linewidth}{ccc|cccc|cccc|cccc}
			\hline
			\multirow{2}*{Method} & \multirow{2}*{Size} & \multirow{2}*{Backbone}& \multicolumn{2}{c}{Occluded-Duke} & \multicolumn{2}{c}{Occluded-REID} \vline &\multicolumn{2}{c}{Partial-REID} & \multicolumn{2}{c}{Partial-iLIDS} \vline &\multicolumn{2}{c}{ Market-1501} & \multicolumn{2}{c}{DukeMTMC-reID} \\ 
			~&~&~&Rank-1 & mAP & Rank-1 & mAP&Rank-1 & Rank-3 & Rank-1 & Rank-3&Rank-1 & mAP & Rank-1 & mAP\\ \hline\hline
			SFR~\cite{he2018recognizing}&$256\times128$& FCN  & 42.3 & 32.0 & - & - &56.9 & 78.5 & 63.9 & 74.8 &-&-&84.83&71.24\\
			FPR~\cite{he2019foreground}&$384\times128$& FCN & - & - & 78.3 & 68.0 & 81.0 & - & 68.1 & -  & 95.4 & 86.6 & 88.6 & 78.4 \\
			PCB~\cite{sun2018beyond}& $384\times128$ & ResNet50 & 42.6 & 33.7& 41.3 & 38.9 &-&-&-&- & 92.3 & 77.4 & 81.8 & 66.1\\
			AMC+SWM~\cite{zheng2015partial}&$128\times48$& ResNet50 & - & - & 31.2 & 27.3 & 37.3 & 46.0 & 21.0 & 32.8 &-&-&-&-\\
			PVPM~\cite{gao2020pose}&$384\times128$& ResNet50 & 47 &37.7&70.4& 61.2 & 78.3 &87.7&-& - &-&-&-&-\\
			PGFA~\cite{miao2019pose}&$256\times128$& ResNet50 & 51.4 & 37.3 & - & - &68.0 & 80.0 & 69.1 & 80.9 &91.2 & 76.8 & 82.6 & 65.5\\
			VPM~\cite{sun2019perceive}&$256\times128$& ResNet50 &-&-&-&- & 67.7&81.9&65.5&74.8 &93.0 &80.8&83.6&72.6\\
			HOReID~\cite{wang2020high}&$256\times128$& ResNet50 & 55.1 & 43.8 & 80.3 & 70.2 &85.3 & 91.0 & 72.6 & 86.4 & 94.2 & 84.9 & 86.9 & 75.6\\
			ISP~\cite{zhu2020identity}&$256\times128$&ResNet50 & 62.8 & 52.3 & - & - &-&-&-&-&95.3&88.6&\textbf{89.6}&80.0\\
			PAT~\cite{li2021diverse}&$256\times128$&ResNet50 & 64.5 & 53.6 & 81.6 & 72.1 &88.0 & 92.3 & 76.5 & 88.2& \textbf{95.4} & 88.0 & 88.8 & 78.2\\ \hline
			\textbf{Ours}&$256\times128$&ResNet50&\textbf{66.2} & \textbf{58.8} & \textbf{83.2} & \textbf{73.5} & \textbf{88.6} & \textbf{92.9} & \textbf{76.9} & \textbf{89.1}  & 95.3 & \textbf{89.4} & 89.3 & \textbf{81.8}\\ 
			\hline
		\end{tabularx}	
		\label{tab:performance}
	\end{table*}
	
	\subsection{Comparison with State-of-the-Art Methods}
	We compare our method with other state-of-the-art models on three person Re-ID benchmarks, including one benchmark for occluded person Re-ID, one benchmark for partial person Re-ID and one benchmark for holistic person Re-ID. In testing, we set all images size in $256 \times 128$.
	
	\textbf{Occluded person Re-ID} We compare our method with other advanced occluded person Re-ID methods on Occluded REID~\cite{miao2019pose} and  Occluded Duke~\cite{miao2019pose}, which are specifically person Re-ID in occlusion situation, demonstrating effectiveness of our method.  The quantitative comparisons with previous models are listed in Tab.~\ref{tab:performance}. From Tab.~\ref{tab:performance} we can observe that our method achieves $66.2\%$ Rank-1 accuracy and $58.8\%$ mAP in Occluded Duke~\cite{miao2019pose}, which surpasses PAT~\cite{li2021diverse}, ISP~\cite{zhu2020identity} by $1.3\%$ and $3.4\%$ in Rank-1 accuracy and $5.2\%$ and $6.5\%$ in mAP. In Occluded REID~\cite{miao2019pose}, our method achieves $83.2\%$ Rank-1 accuracy and $73.5\%$ mAP, which surpasses PAT~\cite{li2021diverse}, HOReID~\cite{wang2020high} by $1.6\%$ and $2.9\%$ in Rank-1 accuracy and $1.4\%$ and $3.3\%$ in mAP. The obtained state-of-the-art performance on the two benchmarks shows the powerful ability of our part segmentation with motion information for occluded person Re-ID.

	\textbf{Partial person Re-ID} To show the ability of our method on partial person Re-ID, we conduct experiments on the Partial REID~\cite{zheng2015partial} and Partial-iLIDS~\cite{zheng2011person}. The result are listed in Tab.~\ref{tab:performance}. We can find that our method still outperforms previous state-of-the-art model proposed by PAT~\cite{li2021diverse} in Rank-1 and Rank-3 accuracy. Our model achieves $88.6\%$ in Rank-1 and $92.9\%$ in Rank-3 in Partial REID, and $76.9\%$ in Rank-1 and $89.1\%$ Rank-3 in Partial-iLIDS, which demonstrate generalization ability of our model for partial person Re-ID.

	\textbf{Holistic person Re-ID} To demonstrate the robustness
 of our model, we test our model on Market-1501~\cite{zheng2015scalable} and DukeMTMC-reID~\cite{ristani2016performance}, which are listed in Tab.~\ref{tab:performance}. For the Market-1501 benchmark, the proposed model has an advantage in mAP by $1.4\%$ and competitive Rank-1 accuracy at the same time. For the DukeMTMC-reID benchmark, the proposed model has improved the mAP up to $81.8\%$, and competitive performance on Rank-1 accuracy.
 

	In conclusion, the obtained state-of-the-art performance on the three benchmarks demonstrates the strong generalization ability and robustness of the proposed method in this work for occluded person Re-ID.

	\section{Conclusion and Future Work}
	\label{sec:conclusion}
	In this paper, we propose a novel motion-aware transformer-based architecture including keypoint detection and part segmentation to address the weaknesses of the existing algorithms for occluded person Re-ID: the limitation of the pre-trained keypoint detector and insufficient fine-grained information for alignment. Adequate experiments have demonstrated that the proposed method efficiently tackles these difficulties and outperforms previous state-of-the-art methods on occluded person datasets in this field. 
	In the future, we are interested in extending this model as a common backbone with high FPS for person Re-ID in realistic scenes.

	{\small

	}
	
\end{document}